\documentclass[conference]{IEEEtran}
\IEEEoverridecommandlockouts

\usepackage{cite}
\usepackage{amsmath,amssymb,amsfonts}
\usepackage{algorithmic}
\usepackage{graphicx}
\usepackage{textcomp}
\usepackage{xcolor}

\usepackage[dvipsnames]{xcolor}

\usepackage{booktabs}
\usepackage{multicol}
\usepackage{multirow}
\usepackage{colortbl}
\usepackage{xcolor}
\usepackage{caption}
\usepackage{tabularray}
\usepackage{rotating}
\usepackage{makecell}
\usepackage{tabularx}
\usepackage{float}
\usepackage{adjustbox}
\usepackage{anyfontsize}
\usepackage{mathtools}
\usepackage{amsmath}
\usepackage{indentfirst}
\usepackage{float}

\usepackage{amssymb}
\usepackage{pifont}
\newcommand{\cmark}{\ding{51}}%
\newcommand{\xmark}{\ding{55}}%
\usepackage{microtype}

    \makeatletter
\def\@fnsymbol#1{\ensuremath{\ifcase#1\or \dagger\or *\or
   \mathsection\or \mathparagraph\or \|\or **\or \dagger\dagger
   \or \ddagger\ddagger \else\@ctrerr\fi}}
    \makeatother

\usepackage{eccvabbrv}

\usepackage{graphicx}
\usepackage{booktabs}

\usepackage[accsupp]{axessibility}  

\usepackage{hyperref}

\usepackage{orcidlink}
\def\BibTeX{{\rm B\kern-.05em{\sc i\kern-.025em b}\kern-.08em
    T\kern-.1667em\lower.7ex\hbox{E}\kern-.125emX}}
\begin{document}

\title{SMamDiff: Spatial Mamba for Stochastic Human Motion Prediction\\
}


\author{Junqiao Fan\IEEEauthorrefmark{1}\thanks{* contribute equally in paper writing.}\orcidlink{0000-0002-8465-5447}\\
\IEEEauthorblockA{\textit{School of Electrical and } \\ \textit{Electronic Engineering, } \\
\textit{Nanyang Technological University.} \\
FANJ0019@e.ntu.edu.sg}
\and
\IEEEauthorblockN{Pengfei Liu\IEEEauthorrefmark{1}\orcidlink{0009-0001-9359-1094}}
\IEEEauthorblockA{\textit{School of Mechanical and} \\ \textit{Aerospace Enginnering,} \\
\textit{Nanyang Technological University.}\\
PENGFEI007@e.ntu.edu.sg}
\and
\IEEEauthorblockN{Haocong Rao\IEEEauthorrefmark{2}\orcidlink{0000-0002-9576-2379}}\thanks{$\dag$ is the corresponding author.}
\IEEEauthorblockA{\textit{School of Computer Science,} \\
\textit{Nanyang Technological University.}\\
HAOCONG001@e.ntu.edu.sg}
}

\maketitle

\begin{abstract}
With intelligent room-side sensing and service robots widely deployed, human motion prediction (HMP) is essential for safe, proactive assistance. However, many existing HMP methods either produce a single, deterministic forecast that ignores uncertainty or rely on probabilistic models that sacrifice kinematic plausibility. Diffusion models improve the accuracy-diversity trade-off but often depend on multi-stage pipelines that are costly for edge deployment. This work focuses on how to ensure spatial-temporal coherence within a single-stage diffusion model for HMP. We introduce SMamDiff, a \textit{S}patial \textit{Mam}ba-based \textit{Diff}usion model with two novel designs: (i) a \textit{residual-DCT} motion encoding that subtracts the last observed pose before a temporal DCT, reducing the first DC component ($f=0$) dominance and highlighting informative higher-frequency cues so the model learns how joints move rather than where they are; and (ii) a \textit{stickman-drawing} spatial-mamba module that processes joints in an ordered, joint-by-joint manner, making later joints condition on earlier ones to induce long-range, cross-joint dependencies. On Human3.6M and HumanEva, these coherence mechanisms deliver state-of-the-art results among single-stage probabilistic HMP methods while using less latency and memory than multi-stage diffusion baselines.
\end{abstract}

\section{Introduction}
\label{sec:introduction}

E-Health is shifting from manual monitoring to intelligent room-side sensing and bringing service robots into daily workflows~\cite{boikanyo2023remote}.
To be genuinely human-centered, these systems are required not only to recognize what a person is doing now, but also to anticipate near-future actions so they can act early, safely, and smoothly in shared spaces~\cite{hoffman2024inferring}.
This motivates human motion prediction (HMP), forecasting likely future poses and trajectories from recent observations to turn passive sensing into proactive assistance~\cite{laplaza2024enhancing}. 
For example, in eldercare and hospital wards, anticipation enables robots and sensing systems to reduce collision risk, coordinate handovers, and prevent falls and injuries.
Potentially, intelligent systems with HMP promise safer, smoother interactions and better patient support. 

Most prior HMP methods~\cite{martinez2017human, li2022skeleton, gao2023decompose} produce a single, deterministic forecast, ignoring the inherently stochastic nature of human movement. Probabilistic HMP methods based on VAEs and GANs introduce diversity, but often yield futures that are overly varied or kinematically unrealistic for clinical and robotic use. Recently, Diffusion Models (DMs) have emerged as a novel generative paradigm for HMP: they learn the motion distribution and sample multiple plausible futures, achieving state-of-the-art (SOTA) kinematic plausibility and realism among probabilistic methods and offering a better accuracy–diversity trade-off. However, to maintain long-range temporal coherence and joint-wise correlations, existing diffusion-based HMP~\cite{barquero2023belfusion} frequently rely on multi-stage pipelines (e.g., pretrained VAE encoders or post-hoc refiners), which increase latency, memory, and engineering complexity, poor-suited for edge deployment on robots and IoT devices. 

Recent work~\cite{chen2023humanmac, tian2024transfusion} explores a single-stage, end-to-end diffusion paradigm for HMP, yet maintaining spatiotemporal coherence within one network and a single training objective remains challenging. To reduce motion jitter and encourage smooth pose evolution over time, these methods commonly encode motion with a discrete cosine transform (DCT) and train DMs in the frequency domain. However, the resulting spectra for human motion representation are highly imbalanced: energy concentrates in the low-frequency band and the DC component ($f=0$) dominates, while high-frequency components ($f>0$) are weak. This bias encourages the model to emphasize average pose/location rather than per-joint dynamics and joint speed. To address the challenges, we first learn human motion in a residual-DCT domain. The design simply subtracts the last frame of the history poses from the human motion sequence, explicitly focusing on how each human joint moves rather than where the pose is located. The proposed residual-DCT representation balances the DC ($f=0$) component with higher-frequency parts, making temporal motion more natural rather than memorizing certain poses. Moreover, for spatial motion coherence between joints, most HMP systems~\cite{sun2024comusion, li2022skeleton} rely on GCN-based spatial modeling, which primarily captures local kinematic adjacency. As a result, long-range dependencies between disconnected joints (e.g., hands and feet) can be overlooked, yielding unnatural running/walking patterns. For better spatial motion coherence learning, we mimic the stickman-drawing method for extracting spatial joint features. During feature extraction, we “draw” joints one by one, making all joints drawn later depend on the already-drawn joints. In this way, we connect all joint features, making their motions mutually dependent

In summary, our contributions are threefold: 
\begin{itemize}
  \item \textbf{Single-stage, end-to-end diffusion in the residual-DCT domain.}
  We present SMamDiff, which encodes motion with a residual-DCT representation, enabling a compact single-stage pipeline that preserves temporal coherence.

  \item \textbf{Stickman-drawing spatial module.}
  An ordered, joint-by-joint spatial feature extraction scheme is designed to capture long-range, cross-joint dependencies beyond local kinematics.

  \item \textbf{SOTA Results.}
  SMamDiff achieves SOTA performance among single-stage probabilistic HMP methods on \emph{Human3.6M} and \emph{HumanEva}, with improved deployment efficiency over multi-stage baselines.
\end{itemize}

\section{Related Work}
\label{sec:related work}
Traditional research on human pose prediction (HMP) has primarily employed deterministic methods to achieve accurate forecasts of single future trajectories~\cite{martinez2017human, li2022skeleton, gao2023decompose}. They typically leverage a regressive paradigm, learning a direct mapping from input historical motion sequences to future motion sequences as output. The recurring neural network (RNN)~\cite{martinez2017human} is a common option to capture temporal dependencies. To capture the motion relationship between the joints, ~\cite{li2022skeleton} used graph convolutional networks (GCNs) to learn spatial representation. However, the high accuracy of these studies usually stems from the relatively poor uncertainty representation of actions in short-term predictions (i.e., single future). As the prediction horizon extends, these deterministic methods often produce averaged motions, resulting in a loss of realism in the predicted futures. 

To overcome this limitation, stochastic methods such as GANs~\cite{barsoum2018hp, gurumurthy2017deligan} and VAEs~\cite{salzmann2022motron,yuan2020dlow,mao2021generating,dang2022diverse} introduce randomness or probabilistic distributions into the modeling of action uncertainty. These methods can generate multiple future trajectories for better motion diversity. Yet, these methods encounter their own bottlenecks, including the need to balance complex loss functions, inherent training instability, and strong reliance on large-scale, high-quality datasets. Moreover, over-emphasizing the diversity of the future also generates unnatural history-future transition, causing a tradeoff between motion diversity and motion realism.

Recently, diffusion-based methods~\cite{wei2023human,barquero2023belfusion,chen2023humanmac,sun2024comusion} have been proposed as a new generative paradigm for stochastic HMP, aiming to balance the generation quality of human pose prediction and the diversity of future trajectories. ~\cite{ahn2023can} proposed a first diffusion framework to predict multiple and diverse futures conditioned on the observed sequence. It is first trained to learn the realistic human motion distribution using a transformer-based network as the motion denoiser. During inference, the history and perturbed future (with Gaussian noises) are concatenated together, and perform step-by-step denoising until realistic motion samples are generated. The diffusion step $t$ is injected as a condition to control the denoising progress. Though achieving promising improvement, generating highly realistic motion transition is still challenging as HMP demands both spatial-temporal coherence, which is complex to model.

To further improve motion realism, MotionDiff~\cite{wei2023human} adopts an encoder-decoder structure. It encodes the history observation into latent features, added with the time embedding $t$ indicating the current diffusion step. To ensure different joint motions are spatially coherent, it employs an additional GCN to refine the predicted actions, which requires a second-stage training. TCD~\cite{saadatnejad2022generic} breaks down the HMP task into short-term and long-term segments, using a strategy of first predicting a few frames close to the observation and then using these short-term results as additional conditions to predict the remaining frames. In addition, TCD also explores the robustness of diffusion-based HMP, such as information loss in the observed frames. It proposes to inject Gaussian noise in the observed frames during training to simulate various missing patterns, thus enhancing the model's robustness. 

Unlike MotionDiff and TCD, which directly learn motion distribution and generate human poses frame-by-frame, BeLFusion adopts a different latent-diffusion strategy. It constructs a latent space using a VAE that enables the model to express and sample diversity at the behavioral level, explicitly decoupling behavior from posture and motion. This design allows both short-term behaviors and long-term actions to be equally encouraged and represented, with diversity no longer constrained by the magnitude of coordinate changes. Moreover, a Behavior Coupler decodes the sampled behaviors into smooth and credible motion continuations, ensuring temporal coherence and sampling realism. However, BelFusion also requires two-stage VAE training and diffusion training. Though it generates realistic samples through multi-stage spatial-temporal modeling, one-stage end-to-end HMP is more desirable for real-world applications.  

Recently, the Discrete Cosine Transform (DCT) has been adopted in HMP~\cite{li2022skeleton} as a powerful tool to model temporal coherence. As human poses are usually smooth and have redundancy in consecutive frames, the human motion can be fully expressed by a few low-frequency components in the frequency domain. Inspired by this, HumanMAC~\cite{chen2023humanmac} and TransFusion~\cite{tian2024transfusion} present a novel perspective on HMP by reformulating the task as a masked completion problem. Instead of relying on multi-stage encoder–decoder pipelines with multiple objectives, the model adopts an end-to-end diffusion framework trained with a single noise prediction loss, which greatly simplifies optimization and reduces reliance on sensitive hyperparameter tuning. During inference, HumanMAC employs a mask-based inpainting strategy that preserves observed motions while progressively refining future sequences, thereby improving temporal consistency and enabling natural transitions across motion categories. Moreover, the use of DCT further enhances efficiency by discarding high-frequency noise and only operating on low-frequency learning, making HMP faster. 

\section{Preliminary}
\subsection{Diffusion Models}
Diffusion models (DMs) are a class of generative models based on stepwise perturbation and iterative denoising. To model complex high-dimensional data distributions that are often intractable (e.g., human motion distribution), DMs construct a gradual noising process that smoothly transforms the original distribution into a simple and analytically tractable distribution (typically a standard Gaussian distribution). Here, we give an overview of DMs, including forward-reverse processes and training-sampling strategies.

\subsubsection{Forward Process}
The forward process is defined as a Markov chain, where each diffusion step $t$ depends only on the previous one $t-1$. It can be expressed as follows.
\begin{equation}
q(x_{0:T}) = q(x_0) \prod_{t=1}^{T} q(x_t \mid x_{t-1}),
\end{equation} where $x_{t}$ is the noisy version of human motion in step $t$.
Specifically, at step $t$, the data is updated as:

\begin{equation}
q(x_t \mid x_{t-1}) = \mathcal{N}\!\left(\sqrt{1-\beta_t}\,x_{t-1}, \; \beta_t I \right),
\end{equation}
where $\beta_t$ is the noise schedule that controls how much noise is injected. Furthermore, a noisy motion sample $x_t$ at any $t$ can be obtained directly from the original data as:
\begin{equation}
x_t = \sqrt{\bar{\alpha}_t} \, x_0 + \sqrt{1-\bar{\alpha}_t}\, \epsilon, 
\quad \epsilon \sim \mathcal{N}(0,I),
\end{equation} where $\alpha_t = 1-\beta_t, \quad \bar{\alpha}_t = \prod_{i=1}^t \alpha_i$.

\subsubsection{Reverse Process}

The goal of the reverse process is to gradually reconstruct the original data from pure noise. Since the forward process is a Gaussian Markov chain, the reverse process also has a Gaussian form, which can be approximated by a neural network $\mu_\theta$ ($\theta$ is the model parameters) as:

\begin{equation}
p_\theta(x_{t-1} \mid x_t) = \mathcal{N}\!\left(x_{t-1} \mid \mu_\theta(x_t,t), \Sigma_\theta(x_t,t)\right).
\end{equation} 
In practice, directly predicting $x_{t-1}$ is difficult and unstable. A more effective approach is to predict the noise $\epsilon$ following DDPM~\cite{ho2020denoising}. In this case, the mean can be rewritten as:
\begin{equation}
\mu_\theta(x_t,t) = \frac{1}{\sqrt{1-\beta_t}}
\left(
x_t - \frac{\beta_t}{\sqrt{1-\bar{\alpha}_t}} \, \epsilon_\theta(x_t,t)
\right),
\end{equation} where $\epsilon_\theta$ is the reformulated network that predicts $\epsilon$.

\subsubsection{Training Objective}
The diffusion model training process essentially minimizes the difference between the true noise and the predicted noise. Therefore, diffusion models typically train a network $\epsilon_\theta(x_t,t)$ to predict noise, optimized using mean squared error:
\begin{equation}
\mathcal{L}(\theta) = \mathbb E_{t,x_0,\epsilon}\!\left[
\left\lVert \epsilon - \epsilon_\theta\!\big(\sqrt{\bar{\alpha}_t}x_0+\sqrt{1-\bar{\alpha}_t}\epsilon,\,t\big)\right\rVert^2
\right]
\end{equation}
This formulation makes training relatively simple, as the model only needs to fit the synthetic noise.  

\subsubsection{Sampling Process}
During inference, the model starts from Gaussian noise $x_T \sim \mathcal{N}(0,I)$ and iteratively updates through multiple denoising steps:
\begin{equation}
x_{t-1} = \frac{1}{\sqrt{1-\beta_t}}
\left(
x_t - \frac{\beta_t}{\sqrt{1-\bar{\alpha}_t}} \epsilon_\theta(x_t,t)
\right)
+ \sigma_t z,
\end{equation}
where $\quad z \sim \mathcal{N}(0,I)$ introduces randomness during reverse sampling. The final result is a sample that belongs to the distribution of the training data. 

While this stepwise denoising of original DDPM ensures high-quality generation, it inevitably incurs a high computational cost, as all $T$ steps must be traversed sequentially. Moreover, naively reducing $T$ often leads to significant degradation in sample quality due to error accumulation. To address these limitations, Denoising Diffusion Implicit Models (DDIM)~\cite{song2020denoising} reformulate the generative process by relaxing the Markov assumption and constructing a non-Markovian inference process that preserves the same marginals $q(x_t|x_0)$ as DDPM. The reverse process is redefined as
\begin{equation}
x_{t-1} = \sqrt{\alpha_{t-1}}x_0 + \sqrt{1-\alpha_{t-1}}\cdot\epsilon_{\theta}(x_t, t),
\end{equation}
where the stochastic noise term can be set to zero, yielding a deterministic mapping from $x_t$ to $x_{t-1}$. Such a non-Markovian formulation allows intermediate steps to be skipped by sub-sampling the time steps, which significantly reduces the number of function evaluations and accelerates generation without compromising perceptual quality. To offer better controllability and ensure historical motion matches the observation during diffusion sampling, we adopt the mask-completion inpainting methods described in HumanMAC~\cite{chen2023humanmac}.

\section{Methodology}
\label{sec:methodology}
\begin{figure*}[!t]
\centering
\includegraphics[width=1.\linewidth]{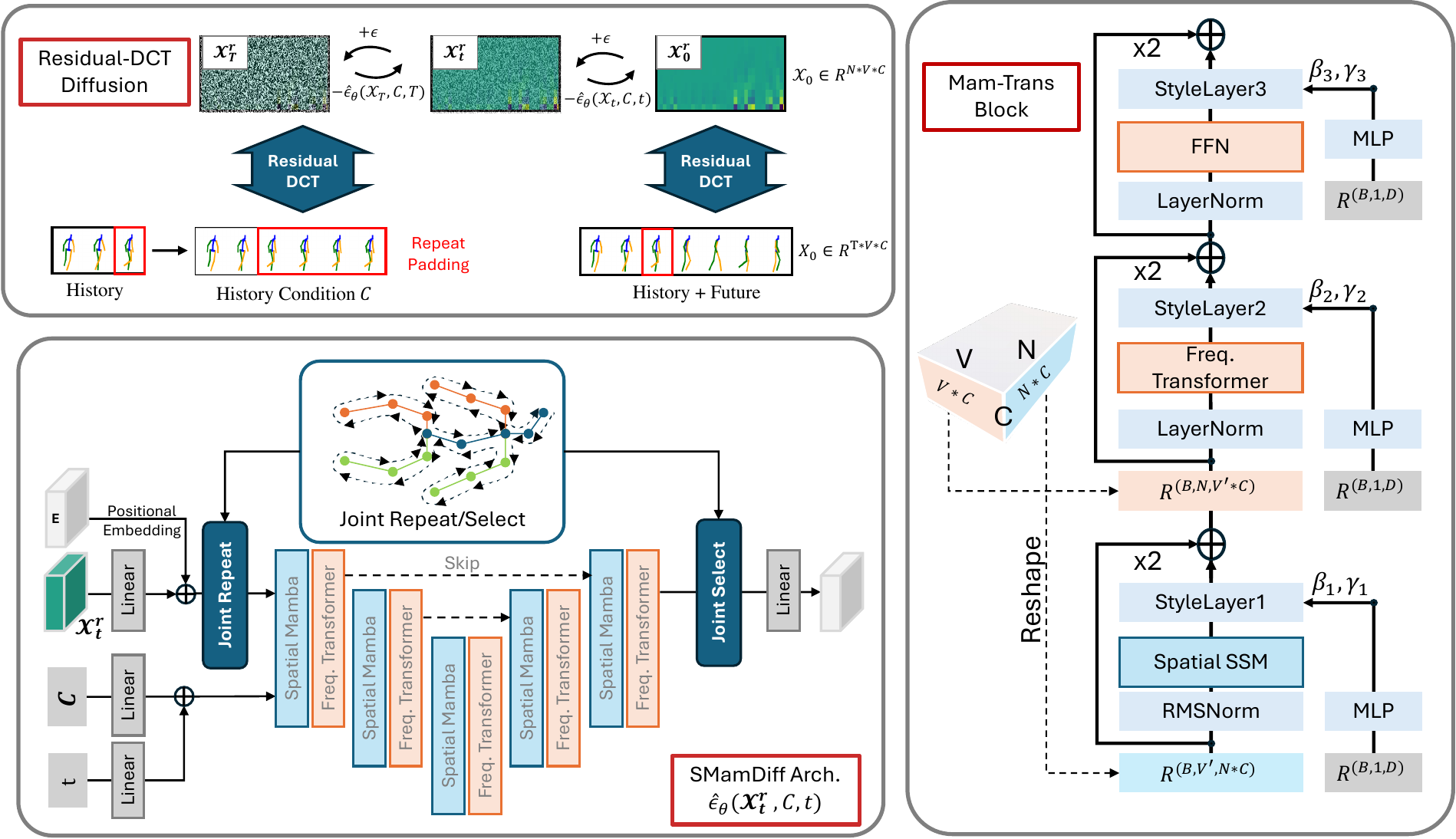}
\caption{System architecture of SMamDiff, the proposed diffusion-based HMP framework. Top-left: overall diffusion process operating in the residual-DCT domain. Bottom-left: network architecture incorporating the spatial Mamba design to mimic stickman drawing. Right: detailed Mam-Trans block used in the SMamDiff architecture.}

\label{system architecture}
\end{figure*}

This work presents a one-stage, diffusion-based HMP framework operating in the frequency domain. As shown in Fig.~\ref{system architecture}, the diffusion model takes $H$ frames of human motion history $X^{[1..H]} = \{x^{[1]}, x^{[2]}, ..., x^{[H]}\}$ as input and outputs the entire sequence comprising both past and future $X^{[1..T]}$, where $T=H+F$ and $F$ denotes the future frame number. Each frame $x^{[i]} \in R^{V\times3}$ is a human skeleton/pose, with $V$ denotes the joint number.

Unlike existing diffusion-based HMP methods, SMamDiff introduces two carefully designed components to enforce spatiotemporal coherence—a key challenge for one-stage diffusion-based HMP. First, to ensure temporal coherence, the model operates in a novel residual-DCT domain, where the diffusion process generates a frequency representation of human motion (see Sec.~\ref{Re-DCT}). Second, we propose a spatial Mamba architecture that mimics the stickman-drawing process to model joint-wise correlations during generation (see Sec.~\ref{spatial-mamba}).


\begin{figure*}[t]
\centering
\includegraphics[width=1.\linewidth]{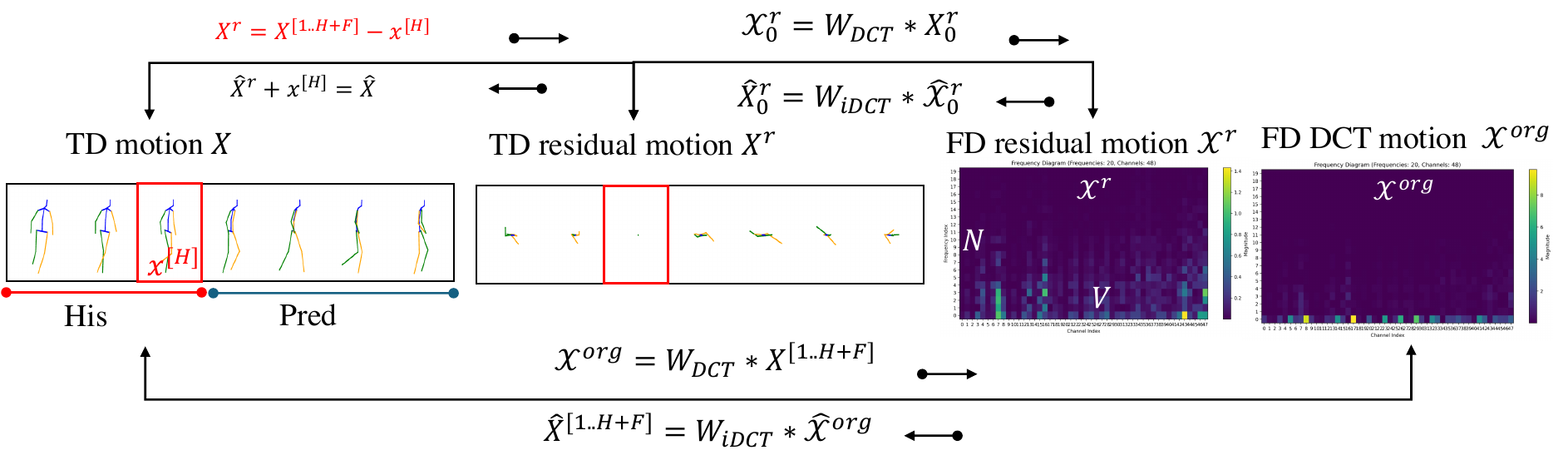}
\caption{Our proposed residual-DCT for converting time-domain (TD) human motion $X$ into frequency-domain (FD) human motion representation $\chi^r$. The traditional FD DCT motion representation $\chi^{org}$ is presented for comparison.}
\label{res-dct}
\end{figure*}

\subsection{Residual DCT Domain Diffusion}
\label{Re-DCT}
As shown in Fig.~\ref{res-dct}, the traditional DCT converts a time-domain (TD) human-motion sequence 
$X^{[1..T]}\!\in\!\mathbb{R}^{T\times 3V}$ into a frequency-domain representation 
$\chi^{\text{org}}\!\in\!\mathbb{R}^{T\times 3V}$ by multiplying the DCT-orthonormal matrix 
$W_{\text{DCT}}$ along the temporal dimension, i.e., $\chi^{\text{org}} = W_{\text{DCT}} X$.
This yields a large energy response at the DC component ($f=0$), since 
$\chi^{[0]} \propto \tfrac{1}{T}\sum_{t=1}^{T} x^{[t]}$ is proportional to the mean joint location
over the sequence. In practice, $\chi^{[0]}$ is often $10\!\times$–$100\!\times$ larger than higher-frequency
coefficients (e.g., $\chi^{[3]}, \chi^{[4]}, \ldots$) that encode velocity/acceleration cues. 
Consequently, learning in the DCT domain can become biased toward average pose/location and may underweight 
fine-grained joint motion, degrading realism.

To alleviate this magnitude-imbalance, we introduce the \emph{residual-DCT} representation.
First, we construct a TD residual sequence $X^{r} = X - \mathbf{1}\,{x^{[H]}}^{\!\top}$ by subtracting the
last observed (history) pose $x^{[H]}$ from every frame ($\mathbf{1}\in\mathbb{R}^{T}$ is an all-ones column).
This focuses the representation on \emph{how} each joint moves rather than \emph{where} it is located, emphasizing
velocity/acceleration patterns. We then apply the temporal DCT to obtain $\chi^{r} = W_{\text{DCT}} X^{r}$.
Notably, for an orthonormal DCT, subtracting a constant affects only the $f=0$ DC term: 
$\chi^{r,[k]} = \chi^{\text{org},[k]}$ for $k>0$, while $\chi^{r,[0]}$ is significantly reduced relative to 
$\chi^{\text{org},[0]}$. This rebalances the spectrum and mitigates DC dominance, as shown in Fig.~\ref{res-dct}. 
Reconstruction to TD from the proposed residual-DCT domain is also lossless: we recover $X$ by $X=W_{iDCT}\chi^{r}+\mathbf{1}\,{x^{[H]}}^{\!\top}$.

Our diffusion model operates directly on $\chi^{r}$. During training, the ground-truth TD motion $X_{0}$ is
converted to $\chi^{r}_{0}$ and the network $\hat{\epsilon}_{\theta}$ learns the human motion distribution in the residual-DCT domain. 
The condition $C$ for the diffusion model is built by repeated-padding the history $X^{[1..H]}$ with $F$ frames of $x^{[H]}$, followed by the
forward residual-DCT transform. At inference, we start from Gaussian noise $\chi^{r}_{T}$ and iteratively denoise with
$\hat{\epsilon}_{\theta}$ to obtain $\hat{\chi}^{r}_{0}$. Finally, we apply the inverse residual-DCT to recover the TD
prediction $\hat{X}_{0}=\mathrm{iDCT}(\hat{\chi}^{r}_{0})+\mathbf{1}\,{x^{[H]}}^{\!\top}$.

\begin{figure}[!t]
\centering
\includegraphics[width=1.\linewidth]{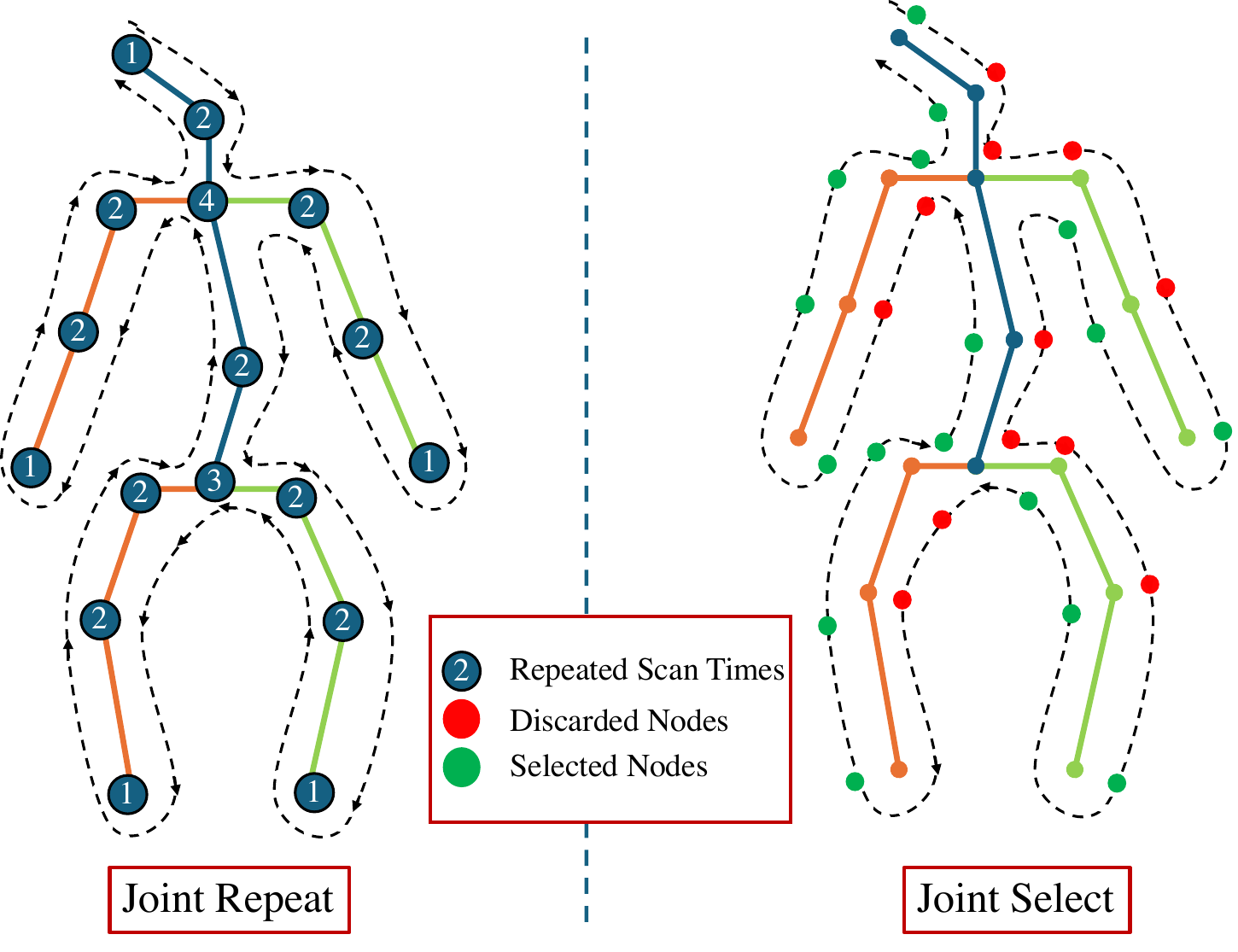}
\caption{Left: “Stickman-drawing” scan starting from the head. The blue number on each joint shows its Joint Repeat
  (how many times that joint is scanned during the forward pass).
  Right: For each joint v, only the last visit is kept as its Joint Select feature; earlier visits are discarded.
  Backtracking at leaves lets later joints condition on earlier ones, creating long-range, cross-joint coherence across all V joints in a single pass.}
\label{spa-mamba} 
\end{figure}

\subsection{Repeated Drawing with Spatial Mamba}
\label{spatial-mamba}

We first model the skeleton as a graph with $V$ joints (nodes) and an adjacency matrix
$\mathbf{A}\in\{0,1\}^{V\times V}$ that encodes the kinematic connections. Instead of using a GCN to
assign a feature to each joint, we argue that the feature should reflect information from all joints,
not only its immediate neighbors. To achieve this, we mimic the ``stickman drawing'' process and
traverse all joints in a fixed, graph-respecting order. The algorithm starts from the head node,
visits adjacent nodes, and returns to parents as needed, drawing joint features node by node. When
drawing a given joint, the model considers all information from previously drawn joints to ensure
consistency.

This motivates us to use an RNN-style model, specifically a state-space model (SSM) suitable for the
sequential drawing paradigm. We implement it with Mamba~\cite{gu2023mamba} as a parallelized, state-of-the-art SSM
architecture. Let $x_v$ be the input at step $v$, $h_v$ the hidden state, and $y_v$ the output. We
parameterize the discrete SSM matrices as functions of $x_v$ via a per-step step size $\Delta_v$:
\begin{equation}
\begin{aligned}
\Delta_v&=\mathrm{softplus}\!\big(w_\Delta^\top x_v\big),\quad
A_v=\exp\!\big(\Delta_v\,\bar A\big),\\
B_v&=(A_v-I)\,\bar A^{-1}\bar B,\quad
C_v=C,
\end{aligned}
\end{equation}
The state update and readout are:
\begin{equation}
\begin{aligned}
h_v &= A_v\,h_{v-1} + B_v\,x_v,\\
y_v &= C_v\,h_v + D_v\,x_v,\quad h_0=\mathbf{0}.
\end{aligned}
\end{equation}
Here, $\bar A$ is a predefined parameter, typically diagonal with negative entries. $\bar B$, $C$, and $D(\cdot)$ are trainable parameters. $w_\Delta$ is a learned vector. $\exp(\cdot)$ denotes the matrix
exponential. $I$ is the identity matrix and $\mathbf{0}$ is the zero vector. In our spatial-Mamba
feature extraction paradigm, at step $v$, $x_v$ is the node feature from the previous layer for the
joint being currently visited; the state $h_v$ accumulates long-range information from all previously visited joints through the input-dependent $A_v,B_v$. When we later visit another joint, its feature is
produced from a state that already summarizes what has been seen elsewhere on the body.

To ensure spatial consistency, we carefully design a scan order comprising \textbf{Joint Repeat} and \textbf{Joint Select} to enable scanning with repetition and backtracking. If each joint were visited only once, the sequence would often ``jump'' after a leaf (for example, from a hand to a distant limb), causing
abrupt context changes in $h_v$. Instead, when we reach a leaf, we go back to its parent (even if it
has already been visited) and continue to the next branch. This creates per-joint repeat counts (as in Fig.~\ref{spa-mamba}), and more joints are integrated into the state for later scans. We keep the final
visit as the joint output, which stabilizes features and strengthens global joint cooperation. See
Fig.~\ref{spa-mamba} for visual explanation.

Finally, we present our block architectural design as the mam-Trans block in Fig.~\ref{system architecture}. Given input features of shape $(B, N, V, C)$ (batch, number of frequency components, joints after repetition, channels), we first merge $N\times C$ per joint and run the spatial-Mamba scan along the joint axis $V$ to produce spatially coherent joint features. We then reshape to merge $V\times C$ per frequency and apply a
frequency Transformer (self-attention along $N$) to model dependencies across frequency components.
Each mam-Trans block has inner residual connections, and multiple blocks are connected with U-Net–style
skips. This ordering enforces spatial coherence first (via the SSM), then refines frequency-wise structure with attention.

\subsection{K-Diversity Training Objective}
During training, we encourage sample-level diversity by producing $K$ parallel noise predictions and backpropagating only through the best-matching one. For $\epsilon\sim\mathcal{N}(0,I)$ and $x_t=\sqrt{\bar{\alpha}_t}\,x_0+\sqrt{1-\bar{\alpha}_t}\,\epsilon$, the model outputs $\widehat{\epsilon}_{\theta,k}(x_t,t,c)$ for $k=1,\ldots,K$. Here is the proposed K-diversity loss:
\begin{equation}
\mathcal{L}_{\text{K-div}}(\theta)
= \min_{k\in\{1,\dots,K\}} \big\|\epsilon-\widehat{\epsilon}_{\theta,k}(x_t,t,c)\big\|_2^2
= \ell_{k^\star},
\end{equation}
where $\ell_k=\|\epsilon-\widehat{\epsilon}_{\theta,k}(x_t,t,c)\|_2^2$ is the per-sample loss and $k^\star=\arg\min_k \ell_k$ is the best index. The gradients are applied only to the best sample $k^\star$. Since only the closest hypothesis is penalized, the remaining $K-1$ heads receive zero gradient on that sample and can specialize to alternative modes, mitigating mode collapse.

\begin{table*}[t]
\centering
\caption{Quantitative results with best-of-many stratergy on Human3.6M and HumanEva-I. Bolded numbers indicate the best results. For all accuracy metrics except for APD, lower values are preferred. The symbol ‘/’ indicates that the results are not reported in the baseline work. We use S, R, and K to indicate the proposed Spatial-mamba module, Residual-DCT module, and K-diversity module, respectively.}

\renewcommand{\arraystretch}{1.5}
\resizebox{1.\textwidth}{!}{


    \begin{tabular}{c|ccc|cccccccc|ccccc}
    \toprule
    \multirow{3}[4]{*}{Type} & \multirow{3}[4]{*}{Method} & \multirow{3}[4]{*}{One-Stage} & \multirow{3}[4]{*}{Loss} & \multicolumn{8}{c|}{Human 3.6}                                & \multicolumn{5}{c}{HumanEva-I} \\
\cmidrule{5-17}          &       &       &       & \multicolumn{2}{c}{Diversity} & \multicolumn{2}{c}{Precision} & \multicolumn{2}{c}{MultiModal GT} & \multicolumn{2}{c|}{Realism} & Diversity & \multicolumn{2}{c}{Precision} & \multicolumn{2}{c}{MultiModal GT} \\
          &       &       &       & APD\textuparrow   & APDE\textdownarrow  & ADE\textdownarrow   & FDE\textdownarrow   & MMADE\textdownarrow & MMFDE\textdownarrow & FID\textdownarrow   & CMD\textdownarrow   & APD\textuparrow   & ADE\textdownarrow   & FDE\textdownarrow   & MMADE\textdownarrow & MMFDE\textdownarrow \\
    \midrule
    \midrule
    \multirow{2}[1]{*}{GAN} & HP-GAN~\cite{barsoum2018hp} & \cmark     & /     & 7.214 & /     & 0.858 & 0.867 & 0.847 & 0.858 & /     & /     & 1.139 & 0.772 & 0.749 & 0.776 & 0.769 \\
          & DeLiGAN~\cite{gurumurthy2017deligan} & \cmark     & 1     & 6.509 & /     & 0.483 & 0.534 & 0.520 & 0.545 & /     & /     & 2.177 & 0.306 & 0.322 & 0.385 & 0.371 \\
    \multirow{4}[1]{*}{VAE} & Motron~\cite{salzmann2022motron} & \cmark     & /     & 7.168 & 2.583 & 0.375 & 0.488 & 0.509 & 0.539 & 13.743 & 40.796 & /     & /     & /     & /     & / \\
          & DLow~\cite{yuan2020dlow}  & \xmark     & 3     & 11.741 & 3.781 & 0.425 & 0.518 & 0.495 & 0.531 & 1.255 & 4.927 & 4.855 & 0.233 & 0.244 & 0.343 & 0.331 \\
          & GSPS~\cite{mao2021generating}  & \xmark     & 5     & 14.757 & 6.749 & 0.389 & 0.496 & 0.476 & 0.525 & 2.103 & 10.758 & 5.825 & 0.233 & 0.244 & 0.343 & 0.331 \\
          & DivSamp~\cite{dang2022diverse} & \xmark     & 3     & 15.310 & 7.479 & 0.370 & 0.485 & 0.475 & 0.516 & 2.083 & 11.692 & 6.109 & 0.220 & 0.234 & 0.342 & \textbf{0.316} \\
    \midrule
    \multirow{4}[2]{*}{DM} & MotionDiff~\cite{wei2023human} & \xmark     & 4     & \textbf{15.353} & /     & 0.411 & 0.509 & 0.508 & 0.536 & /     & /     & 5.931 & 0.232 & 0.236 & 0.352 & 0.320 \\
          & BeLFusion~\cite{barquero2023belfusion} & \xmark     & 4     & 7.602 & 1.662 & 0.372 & 0.474 & \textbf{0.473} & 0.507 & 0.209 & 5.988 & /     & /     & /     & /     & / \\
          & HumanMAC~\cite{chen2023humanmac} & \cmark     & 4     & 6.301 & /     & 0.369 & 0.480 & 0.509 & 0.545 & /     & /     & \textbf{6.554} & 0.209 & 0.223 & \textbf{0.342} & 0.335 \\
          & CoMusion~\cite{sun2024comusion} & \cmark     & 1     & 7.632 & \textbf{1.609} & 0.350 & 0.458 & 0.494 & \textbf{0.506} & 0.102 & 3.202 & /     & /     & /     & /     & / \\
    \midrule
    \multirow{3}[2]{*}{Ours} & SMamDiff(S)     & \cmark     & 1     &   /    &    /   &    /   &  /     &   /    &  /     &   /    & /      & 5.577 & 0.198 & 0.213 & /     & / \\
          & SMamDiff(R + S) & \cmark     & 1     & 5.515 & 2.816 & 0.346 & 0.463 & 0.509 & 0.542 & \textbf{0.0514} & 3.666 & 1.858 & 0.193 & 0.233 & /     & / \\
          & SMamDiff(K + R + S) & \cmark     & 1     & 7.220 & 1.789 & \textbf{0.345} & \textbf{0.457} & 0.500 & 0.525 & 0.064 & \textbf{2.324} & 2.540 & \textbf{0.187} & \textbf{0.205} & 0.344 & 0.340 \\
    \bottomrule
    \end{tabular}%

}
\label{tab:main}
\end{table*}

\begin{figure*}[t]
\centering
\includegraphics[width=1.\linewidth]{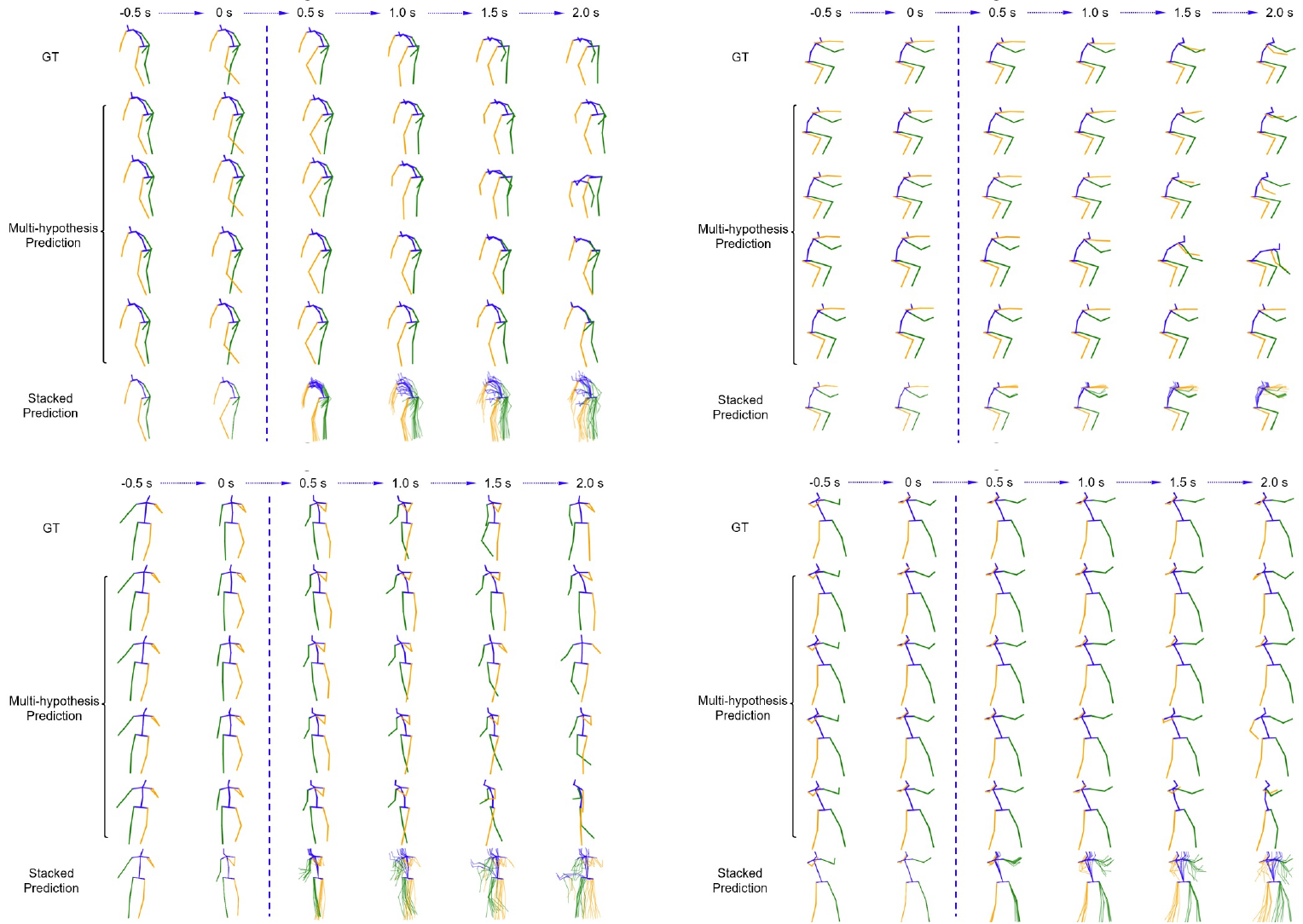}
\caption{Qualitative comparisons by visualizing prediction results for 4 actions from the Human3.6M dataset. Each column depicts the temporal evolution of motion from $0.5$~s to $2.0$~s, with the ground-truth sequence shown at the top, followed by the multi-hypothesis predictions and a stacked visualization of all predicted trajectories.}
\label{qual}
\end{figure*}

\section{Experiments}
\label{sec:experiments}

\subsection{Experiment Setup}
\textbf{Datasets.}
In our experiments, we adopt two widely-used human motion prediction benchmarks: Human3.6M~\cite{ionescu2013human3} and HumanEva-I~\cite{sigal2010humaneva}. Human3.6M is one of the largest motion capture datasets, consisting of approximately 3.6 million frames, 17 action categories, and 11 subjects recorded at 50 Hz. Following the standard protocol, we use a 17-joint skeleton instead of the original 32 joints, train on 5 subjects (S1, S5, S6, S7, S8), and test on 2 subjects (S9, S11). Each sequence uses 25 frames (0.5 s) as the observation to forecast the following 100 frames (2 s). HumanEva-I is a relatively smaller dataset with 4 subjects performing 6 common actions, recorded at 60 Hz. We follow the official train/test split, using 15 observed frames (0.25 s) to predict the next 60 frames (1 s). This setup ensures fair comparison with previous works.

\textbf{Evaluation metrics.}
We adopt a comprehensive evaluation protocol covering four key aspects:

\begin{itemize}
    \item \textbf{Diversity.} We report Average Pairwise Distance (APD), which measures the overall spread among generated motion samples (higher is better), and Average Pairwise Distance Error (APDE), which evaluates the average error across all sample pairs (lower is better). These metrics capture the variability of generated motions.

    \item \textbf{Precision.} To assess prediction accuracy, we compute Average Displacement Error (ADE), the mean L2 distance across all time steps, and Final Displacement Error (FDE), the L2 distance at the last predicted frame. Lower values indicate higher precision.

    \item \textbf{Multi-Modal Ground Truth.} In settings where multiple plausible futures exist, we use Multi-Modal ADE (MMADE) and Multi-Modal FDE (MMFDE), which match predictions against the closest ground-truth mode, providing a fairer evaluation under uncertainty.

    \item \textbf{Realism.} Finally, we report perceptual metrics including Fréchet Inception Distance (FID) and Chamfer Motion Distance (CMD), which measure the distributional similarity and geometric consistency between predicted motions and the real motion trajectories, reflecting visual plausibility and naturalness.
\end{itemize}

This categorization allows a more systematic analysis, jointly considering diversity, accuracy, multi-modal coverage, and realism.

\textbf{Baseline.}
To validate the effectiveness of our method, we conduct a comparative study with several representative state-of-the-art approaches. 
Specifically, we compare against HP-GAN~\cite{barsoum2018hp}, DeLiGAN~\cite{gurumurthy2017deligan}, Motron~\cite{salzmann2022motron}, DLow~\cite{yuan2020dlow}, GSPS~\cite{mao2021generating}, DivSamp~\cite{dang2022diverse}, MotionDiff~\cite{wei2023human}, BeLFusion~\cite{barquero2023belfusion}, HumanMAC~\cite{chen2023humanmac}, and CoMusion~\cite{sun2024comusion}. 
These methods cover a wide spectrum of generative paradigms, including GAN-based, VAE-based, diffusion-based, and sampling-based frameworks, 
providing a comprehensive benchmark for evaluating the accuracy, diversity, and realism of motion prediction models.

\textbf{Implementation Details.}
We configure SMamDiff with 1,000 diffusion steps and 100 DDIM sampling steps, using a cosine noise schedule. We train for 1,000 epochs on Human3.6M and 500 epochs on HumanEva-I with a batch size of 64. We use Adam with a learning rate of 3e-4 and reduce the learning rate by a factor of 0.8 every 100 epochs. During training, we apply classifier-free guidance by dropping the history condition with probability 0.2.

The network uses 4 mam-trans blocks with a hidden size of 384. Its parameter count is significantly smaller than the HumanMAC baseline~\cite{chen2023humanmac}, which uses 8 Transformer layers with hidden size 512. For both DCT and residual-DCT variants, we keep the lowest 20 frequency coefficients to represent motion. We set $K = 5$ for the K-diversity training objective. All experiments are implemented in PyTorch and run on a single NVIDIA RTX 3090 GPU.

\subsection{Comparison with the State-of-the-Arts}

\textbf{ Quantitative Results.}
We conduct a comprehensive comparison of our proposed SMamDiff with representative human motion prediction models from various paradigms, and report the quantitative results in Table~\ref{tab:main}. Consistent with previous findings, diffusion-based models (DM) generally outperform GAN-based and VAE-based approaches across most evaluation metrics, underscoring their superior ability to capture the complex distribution of human motion.

Relative to the baseline HumanMAC, SMamDiff consistently improves on all metrics except MMADE, indicating that our model not only generates more accurate predictions but also produces trajectories that are perceptually more realistic, all while maintaining competitive diversity. These results highlight SM’s ability to produce motion sequences that remain faithful to the ground truth distribution.

When compared with the state-of-the-art CoMusion, SMamDiff attains slightly lower scores on the multi-modal ground-truth metrics (MMADE and MMFDE), which is expected since SMamDiff is specifically designed to emphasize prediction accuracy rather than explicitly modeling multiple plausible futures. Nevertheless, SMamDiff achieves notable gains on the realism metrics, delivering 37.3\% improvement in FID and 27.4\% in CMD, suggesting that the motions generated by SMamDiff are visually smoother, more natural, and better aligned with physical constraints.

Most importantly, SMamDiff achieves the best performance on accuracy-oriented metrics (ADE and FDE) among all compared methods, confirming that our approach produces temporally consistent motion trajectories that remain closer to the ground truth across the entire prediction horizon. Regarding diversity, SMamDiff yields moderate APD and APDE values, achieving a desirable trade-off between diversity and plausibility. In line with prior observations in~\cite{sun2024comusion}, we argue that excessive diversity may lead to unrealistic or physically implausible motions. SMamDiff instead maintains a balanced level of diversity, ensuring behavioral realism and stability in the generated predictions.

\textbf{ Qualitative Results.}
Overall, the motions predicted by SMamDiff remain well aligned with the ground-truth trajectories, producing smooth, temporally coherent, and physically plausible sequences over the entire prediction horizon. The multi-hypothesis results exhibit moderate yet meaningful diversity without introducing unrealistic limb poses or sudden discontinuities. For instance, in Fig.~\ref{qual}, SMamDiff accurately preserves the global body orientation and predicts leg swing motions that are natural and stable.

Similarly, SMamDiff maintains consistent joint coordination and generates natural transitions between frames, leading to realistic action portrayals. In examples, SMamDiff successfully models both global translation and fine-grained limb dynamics, yielding stable predictions even in the long-term horizon beyond 1.5~s, where many approaches typically accumulate substantial errors.

These qualitative observations corroborate our quantitative findings, particularly the improvements in FID and CMD, confirming that SMamDiff produces motion sequences that are not only more accurate but also visually realistic and behaviorally consistent, while retaining an appropriate level of diversity across multiple hypotheses.

\section{Conclusion}\label{sec:conclusion}
We present a one-stage, end-to-end diffusion framework for stochastic human motion prediction. To address the spatial–temporal coherence challenges inherent to single-stage, single-loss HMP, we (i) train in a residual-DCT domain to balance magnitudes across low- and high-frequency components, and (ii) introduce a spatial-Mamba module that mimics a “stickman-drawing” process to enforce joint-wise spatial coherence. Our method attains state-of-the-art accuracy among single-stage HMP approaches and delivers notably more realistic motions. The one-stage design also yields strong efficiency and practical implementation benefits for E-health and robotics deployment. A remaining limitation is sample diversity, which we identify as an important direction for future work.

%
%
\bibliographystyle{splncs04}


\end{document}